\documentclass[lettersize,journal]{IEEEtran}
\usepackage[colorlinks,urlcolor=blue,linkcolor=blue,citecolor=blue]{hyperref}

\usepackage{graphicx}
\usepackage{amsmath}
\usepackage{amssymb}
\usepackage{algorithm}
\usepackage{algpseudocode}
\usepackage{booktabs}
\usepackage{caption}
\usepackage{subcaption}

\usepackage{array}
\usepackage{multirow}
\usepackage{float}
\usepackage[table]{xcolor}
\usepackage{mathtools}
\usepackage{adjustbox}

\usepackage{ tipa }

\def\BibTeX{{\rm B\kern-.05em{\sc i\kern-.025em b}\kern-.08em
    T\kern-.1667em\lower.7ex\hbox{E}\kern-.125emX}}

\begin{document}

\title{Reinforcing Competitive Multi-Agents for Playing {\textquoteleft{So Long Sucker}\textquoteright}}

% \title{Designing Competitive Multi-Agents for Playing {\textquoteleft{So Long Sucker}\textquoteright} Using Deep Reinforcement Learning}

% Learning to Play \textquoteleft{So Long Sucker}\textquoteright: Autonomous Agents in Multi-Agent Reinforcement Learning

\author{Medant Sharan, 
    Chandranath Adak,~\IEEEmembership{Senior Member,~IEEE}
\thanks{Medant Sharan is with the Dept. of Informatics, King's College London, WC2R 2LS United Kingdom.\\
Chandranath Adak is with the Dept. of CSE, Indian Institute of Technology Patna, Bihar 801106, India. 
Corresponding author: C. Adak (email: chandranath@iitp.ac.in)
}
}

% \author{Anonymous
% \thanks{Anonymous}
% }

% \markboth{IEEE Transactions on Games}{}
% {Medant \MakeLowercase{\textit{et al.}}: xxx}

% \markboth{Medant \MakeLowercase{\textit{et al.}}}{}

\maketitle

\begin{abstract}
This paper investigates the strategy game \emph{So Long Sucker} (SLS) as a novel benchmark for multi-agent reinforcement learning (MARL). Unlike traditional board or video game testbeds, SLS is distinguished by its coalition formation, strategic deception, and dynamic elimination rules, making it a uniquely challenging environment for autonomous agents. We introduce the first publicly available computational framework for SLS, complete with a graphical user interface and benchmarking support for reinforcement learning algorithms. Using classical deep reinforcement learning methods (e.g., DQN, DDQN, and Dueling DQN), we train self-playing agents to learn the rules and basic strategies of SLS. Experimental results demonstrate that, although these agents achieve roughly half of the maximum attainable reward and consistently outperform random baselines, they require long training horizons ($\sim$2000 games) and still commit occasional illegal moves, highlighting both the promise and limitations of classical reinforcement learning. Our findings establish SLS as a negotiation-aware benchmark for MARL, opening avenues for future research that integrates game-theoretic reasoning, coalition-aware strategies, and advanced reinforcement learning architectures to better capture the social and adversarial dynamics of complex multi-agent games.
% \footnote{At  \href{https://anonymous.4open.science/r/SLSGameCode}{https://anonymous.4open.science/r/SLSGameCode}, initial codes can be found. The full SLS environment will be published on acceptance of the paper.}
%
% This paper introduces \emph{So Long Sucker} (SLS), a negotiation- and betrayal-driven strategy game, as a novel benchmark for multi-agent reinforcement learning (MARL). Unlike existing testbeds that emphasize either cooperation (coordination tasks) or pure competition (e.g., Go, StarCraft II), SLS uniquely integrates coalition formation, shifting alliances, and inevitable betrayal. We present the first open-source computational framework for SLS, complete with a graphical user interface (GUI) and reproducible benchmarking tools. Using classical deep reinforcement learning baselines (DQN, DDQN, and Dueling DQN), we show that agents learn legal play and achieve roughly half of the maximum attainable reward, yet still commit occasional illegal moves and require long training horizons ($\sim$2000 games). These results highlight both the promise and limitations of classical DRL in socially complex domains. By formalizing SLS as a reproducible MARL testbed, this work establishes a foundation for future research that combines deep learning with game-theoretic reasoning to capture the rich coalition and negotiation dynamics of human strategic play.

\end{abstract}

% \begin{IEEEkeywords}
% Deep Reinforcement Learning, DQN, DDQN, Dueling DQN, Multi-Agent Learning, Game Theory, \textit{So Long Sucker} (SLS), Strategy Games, Adversarial Learning, Coalition Formation, Diplomacy, Rule Learning, Multi-Agent Systems, Social Dynamics, Game AI
% \end{IEEEkeywords}


\section{Introduction}
\label{sec:intro}

\IEEEPARstart{S}{o Long Sucker} (SLS) is a four-player strategy game devised by Hausner, Nash, Shapley, and Shubik \cite{nash} as a playful yet profound exploration of game-theoretic principles. Unlike many traditional board or video games that emphasize tactical positioning, SLS revolves around coalition formation, temporary alliances, and inevitable betrayal. Victory depends not only on legal play but also on diplomacy and opportunism, making SLS a distinctive testbed for studying trust, negotiation, and social dilemmas. While human players typically grasp its mechanics within a few rounds, sustained success requires adapting to adversarial strategies and shifting alliances.

Despite this richness, SLS has received little attention in computational game AI. Most multi-agent reinforcement learning (MARL) benchmarks emphasize either cooperative objectives (e.g., coordination tasks) or adversarial competition (e.g., two-player zero-sum games). Breakthroughs such as Go \cite{alphago}, StarCraft II \cite{vinyals2019grandmaster}, and Diplomacy \cite{meta2022diplomacy}, agents demonstrate the power of deep reinforcement learning (DRL) for large-scale strategic reasoning, but they do not fully capture the hybrid dynamics of coalition and betrayal that characterize SLS. This gap limits our understanding of how autonomous agents can learn strategies in environments where optimal play may involve both collaboration and defection \cite{dqn_drl}.

In this paper, we take a first step toward bridging this gap by training DRL agents to play SLS. Specifically, we investigate classical value-based methods, Deep Q-Network (DQN), Double DQN (DDQN), and Dueling DQN, as baselines for learning the rules and fundamental strategies of the game \cite{suttonbook_RL}. To support this research, we implement and publicly release a machine-learning–ready version of SLS, complete with a graphical user interface (GUI) and benchmarking tools. Our framework enables reproducible evaluation and establishes SLS as a negotiation-aware benchmark for future MARL research. 
Our contributions are threefold: 

\textbf{\emph{(i)}} We design and release the first computational framework for SLS, tailored for reinforcement learning research and equipped with a GUI for experimentation.

\textbf{\emph{(ii)}} We benchmark classical DRL algorithms (DQN, DDQN, Dueling DQN) in SLS, analyzing their ability to acquire legal play proficiency and partial strategic awareness.

\textbf{\emph{(iii)}} We establish SLS as a coalition and betrayal aware benchmark for MARL, motivating future research on hybrid approaches that combine DRL with game-theoretic reasoning.

% The remainder of this paper is organized as follows. 
% Section \ref{sec:related_work} discusses the background,  
% Section \ref{3sec:proposed} describes the proposed methodology, 
% Section \ref{4sec:result} presents experimental results, and 
% Section \ref{5sec:conclusion} concludes this paper.

The remainder of this paper is organized as follows: 
Section~\ref{sec:related_work} prepares the work background, Section~\ref{3sec:proposed} details the methodology, Section~\ref{4sec:result} presents the results, and Section~\ref{5sec:conclusion} concludes the paper.

\section{Background}
\label{sec:related_work}

% \emph{So Long Sucker} (SLS)~\cite{nash} is a strategic board game grounded in game theory, designed to highlight concepts such as coalition formation, temporary alliances, and eventual betrayal. In its original form, SLS allows free-form deal-making and negotiations, making the gameplay highly non-sequential and challenging to model computationally. For this study, we adopt a simplified, sequential variant that omits explicit negotiations while retaining the game’s essential adversarial and strategic dynamics. This design aligns more naturally with the requirements of multi-agent reinforcement learning (MARL), ensuring agents can still engage in capture, elimination, and dynamic turn assignment without requiring complex language or contract enforcement.

SLS \cite{nash} is a strategic board game rooted in game theory, originally designed to highlight coalition dynamics and adversarial play. While the original version allows free-form negotiations and deal-making, we adopt a simplified sequential variant that omits explicit negotiations. This adjustment makes the game more tractable for MARL while preserving its core adversarial mechanics, including captures, eliminations, and dynamic turn assignment.

\subsection{Game Variant}\label{sec:game_variant}
Our analysis is based on the \emph{Generalized Hofstra’s version} of SLS~\cite{jerade2024so,decarufel2024longsuckerendgameanalysis}. Four players, each assigned a distinct color (e.g., blue, green, red, yellow), begin with $c \in \mathbb{N}$ chips of their own color. The game proceeds on a board with $k \in \mathbb{N}$ empty piles, where each pile is an ordered stack of chips. The winner is the last surviving player, although interestingly, victory is still possible for a player who has no chips remaining at the moment of elimination. This version also introduces a backtracking rule to handle cases where eliminated players must be skipped until a valid participant is found, a detail overlooked in prior analyses.

Traditionally, the Generalized Hofstra’s version uses a progressive scoring system $\{1,2,3,4\}$, rewarding players based on elimination order. For reinforcement learning, however, we introduce a \emph{Zero-Sum Generalized Hofstra’s version}, where the payoff structure is modified to $\{0,0,0,\text{\textrtails}\}$ with $\text{\textrtails} \in \mathbb{N}^+$. This zero-sum formulation eliminates ambiguity in intermediate scores, providing a clearer reward signal and enabling faster convergence of baseline DRL agents.

\subsection{Rules of Zero-Sum Generalized Hofstra's SLS}
\label{sec:rules}

\subsubsection{Starting the Game}
\begin{itemize}
    \item 
    Each player receives 5 chips of their assigned unique color.
    \item 
    A random player begins by placing a chip onto the playing area and designating the next player.
\end{itemize}

\subsubsection{Gameplay Mechanics}
\begin{itemize}
    \item On a turn, a player places a chip of any color, either starting a new pile or extending an existing pile (maximum 6 active piles).
    \item If no pile is captured, the player must choose the next participant such that their starting color is not present in the pile. If all colors are represented, the next player is the one whose color is deepest in the stack.
    \item A pile is captured when two consecutive chips of the same color appear. The player of that color must kill/ remove one chip from the pile, and the capturing player claims the remaining chips and takes the next turn.
    \item Chips of one color held by another player are treated as prisoners.
    \item 
% A player with no chips available on their turn is eliminated, since negotiation is not a part of our current framework.
A player with no chips available on their turn during gameplay is eliminated, since negotiation is excluded in our current framework. 
    \item If a defeated player’s chip captures a pile, the entire pile is removed, and the turn reverts to the capturing player.
\end{itemize}

% \subsubsection{Order of Play}\label{rulevariation}
% \begin{itemize}
%     \item After a capture, the player whose chip triggered the capture takes the next turn.
%     \item If a player is eliminated, the turn reverts to the player who previously passed to them. This backtracking continues until a valid player is identified;
% \textcolor{red}{if none exists, a random assignment is made.}
% In the original game introduced by Nash et al (and other developing papers) the idea of such a player not existing was not anticipated, making the rules incomplete. We chose to do random assignment in case this situation arose, making the rules complete.

%     \item When no capture or elimination occurs, the last mover designates the next player:
%     \begin{itemize}
%         \item[---] Any player whose color is not already present in the stack may be chosen, including the mover.
%         \item[---] If all colors are represented, the next turn goes to the player whose chip lies deepest in the stack.
%     \end{itemize}
% \end{itemize}

\subsubsection{Order of Play}\label{rulevariation}
\begin{itemize}
    \item After a capture, the player whose chip triggered the capture takes the next turn.
    \item If a player is eliminated, the turn reverts to the player who previously passed to them. This backtracking continues until a valid player is identified. 
    If no such player exists, the turn is assigned randomly. The original formulation in \cite{nash} and subsequent studies \cite{jerade2024so,decarufel2024longsuckerendgameanalysis} did not anticipate this situation, leaving the rules incomplete. 
    % Our modification resolves this ambiguity and ensures completeness of the rules.
    Our modification resolves the ambiguity, ensuring rule completeness.
    \item When no capture or elimination occurs, the last mover designates the next player:
    \begin{itemize}
        \item[---] Any player whose color is not already present in the stack may be chosen, including the mover.
        \item[---] If all colors are represented, the next turn goes to the player whose chip lies deepest in the stack.
    \end{itemize}
\end{itemize}

% \subsubsection{Order of Play}\label{rulevariation}
% \begin{itemize}
%     \item After a capture, the player whose chip triggered the capture takes the next turn.
%     \item If a player is eliminated, the turn reverts to the player who previously passed to them. This backtracking continues until a valid player is identified. 
%     In cases where no such player exists, we assign the turn randomly. Notably, the original formulation by Nash et al.\ did not account for this scenario, leaving the rules incomplete. Our modification resolves this ambiguity and ensures completeness of the game rules.
% \end{itemize}

\subsubsection{Winning Conditions} 
\begin{itemize}
    \item 
    The game ends when only one player remains, who is declared the winner.
    \item 
    A player may still win even with zero chips, provided all others are eliminated.
\end{itemize}

\subsection{SLS as a Benchmark}
Although the original formulation of SLS included free-form deal-making and explicit coalition agreements, we adopt a sequential, zero-sum variant to ensure tractability and reproducibility for MARL. This version preserves the essential adversarial elements, i.e., pile captures, eliminations, prisoner chips, and dynamic turn control; so agents continue to face coalition-like dynamics and opportunities for strategic deception without requiring explicit dialogue.

Recent MARL research has achieved remarkable success in large-scale cooperative and competitive domains such as Go, StarCraft II, and Diplomacy~\cite{alphago,vinyals2019grandmaster,meta2022diplomacy}. However, these benchmarks focus primarily on either pure competition or cooperation, with little emphasis on hybrid dynamics involving shifting alliances and betrayal. Prior work on socially rich settings often depends on explicit communication channels or handcrafted interaction rules~\cite{foerster2016learning}, which limits generality. To the best of our knowledge, SLS has not previously been studied as a computational benchmark. By formalizing a reproducible sequential variant and benchmarking baseline DRL algorithms, this paper positions SLS as a coalition- and betrayal-aware testbed for advancing research in MARL.

\section{Proposed Methodology}
\label{3sec:proposed}

To transform SLS into a machine-learning-ready benchmark, we formalize the game as a MARL environment with explicit state, action, and reward representations. Our design preserves the core adversarial mechanics of SLS—captures, eliminations, prisoner chips, and dynamic turn control, while providing a tractable interface for DRL agents.

% We formalize SLS as a MARL environment with explicit state, action, and reward representations, preserving its core adversarial mechanics—captures, eliminations, prisoner chips, and dynamic turn control, while ensuring a tractable interface for DRL agents.

\subsection{Reinforcement Learning Formulation}
We cast the SLS in reinforcement learning, which is a machine learning paradigm, where an agent interacts with an environment to maximize cumulative reward. Formally, the problem is modeled as a Markov Decision Process (MDP) defined by:
% We transform SLS as a MARL environment, preserving its core adversarial mechanics—captures, eliminations, prisoner chips, and dynamic turn control—while providing a tractable interface for DRL agents. 
% We formalize SLS as an MARL environment, preserving its core adversarial mechanics—captures, eliminations, prisoner chips, and dynamic turn control, while ensuring a tractable interface for DRL agents. 
% Formally, the game is represented as a Markov decision process, where agents interact with the environment through state, action, and reward functions to maximize cumulative return.

\begin{itemize}
    \item \textit{State space} $\mathcal{S}$: The set of all possible environment states, i.e., game states.
    
    \item \textit{Action space} $\mathcal{A}$: The set of all actions available to the agent, i.e., the discrete set of valid moves.
    
    \item \textit{Transition function} $p(s'|s,a)$: This represents the probability of moving from state $s$ to $s'$, when takes action $a$. In deterministic environments, this transition is often represented by a fixed mapping.
    
    \item \textit{Reward function} $r(s,a,s')$: This assigns a scalar value to each transition $(s,a,s')$, quantifying the immediate utility of executing action $a$ in state $s$ and reaching the subsequent state $s'$. 

    % The reward function $R(s, a, s')$ provides a scalar reward signal to the agent based on the action taken at a particular state. The goal of the agent is to maximize the total accumulated reward over time.    
    
    \item \textit{Policy} $\pi(a|s)$: This is an agent’s strategy for selecting an action $a$ given a state $s$. The policy can be deterministic, where a specific action is chosen for each state, or stochastic, where actions are selected with certain probabilities.
    
\end{itemize}

% The objective is to learn a policy $\pi(a|s)$ that maximizes the expected discounted return.

\noindent
The goal is to find an optimal policy $\pi^*$ that maximizes the expected discounted return: 
\begin{equation}
    \mathcal{R}_t = \sum_{k=0}^{\infty} \gamma^k r_{t+k+1}
\end{equation}
where, $\gamma \in [0,1]$ the discount factor, $r_{t+k+1}$ the reward received after future $k$ steps.  Classical algorithms include value-based methods (DQN, DDQN, Dueling DQN), policy-based methods, and actor–critic hybrids.

% and $\pi^*$ the policy that maximizes the expectation of this return.

%=====================================

\subsection{State Representation}

% Each state $s_t$ encapsulates the full game configuration:
% \begin{itemize}
%     \item \textbf{Board configuration:} active piles with chip order.
%     \item \textbf{Chip ownership:} chips held by each player and prisoners.
%     \item \textbf{Elimination status:} flags for dead players.
%     \item \textbf{Turn context:} identity of the current player and game phase.
%     \item \textbf{Step count:} tracking episode progression.
% \end{itemize}
% This encoding balances completeness and compactness, enabling neural networks to process the game efficiently.

%==================

% The state $s_t$ at timestep $t$ captures all relevant information about the current game environment. It is structured as a vector with specific components representing the game board, player chips, eliminated chips, the current player, game phase, and step count:
% \begin{equation}
% \begin{split}
% s_t = \{ \text{Board State}, \text{Player Chips}, \text{Dead Chips}, \\
% \text{Current Player}, \text{Game State}, \text{Steps} \}    
% \end{split}
% \end{equation}

The state $s_t$ at timestep $t$ encodes all information necessary to describe the game environment. 
This encoding balances completeness and compactness, enabling neural networks to process the game efficiently.
It is represented as a structured vector comprising the following components:
\begin{equation} \label{eq:s_t}
\begin{split}
s_t &= \big( \textit{Board Configuration}, \textit{Player Chips}, \textit{Eliminated} \\
    &\quad \textit{Chips}, \textit{Current Player}, \textit{Game Phase}, \textit{Step Count} \big); 
\end{split}
\end{equation}
where, each term corresponds to a distinct component of the state representation, as discussed below:

\subsubsection{Board Configuration}
This encodes the chip distribution across rows and piles, which can be computed as $n_{rows} \times n_{players} \times n_{max\_pile}$; where, 

% $\text{num\_rows} \times \text{num\_players} \times \text{max\_pile\_size}$. This term represents the current state of the game board. 
    \begin{itemize}
        \item $n_{rows}$ is the number of playable rows in the game.
        
        \item $n_{players}$ represents the number of players, each of whom has their own set of chips that can be placed in any row.
        
        % \item \texttt{max\_pile\_size} represents the upper limit on the size of a pile, assuming every player has placed the maximum number of chips in a single row. It is defined as $$n_{players}$ \times \texttt{max\_chips}$, where $\texttt{max\_chips}$ is the maximum number of chips a player can hold. 
        
        \item $n_{max\_pile}$ denotes the upper limit on pile size, corresponding to the case where all players place their maximum chips in a single row. It is given by $n_{players} \times n_{max\_chips}$, where $n_{max\_chips}$ denotes the maximum chips that a player can hold.        
    \end{itemize}
    
    % Therefore, $$n_{rows}$ \times $n_{players}$ \times \texttt{max\_pile\_size}$ captures all possible states of the board, accounting for each row, player, and pile configuration.

\noindent
Therefore, $n_{rows} \times n_{players} \times n_{max\_pile}$ encodes the board configuration by capturing every possible combination of rows, players, and pile sizes.

\subsubsection{Player Chips}
Since each player may hold chips originating from any other player, the information is stored in a matrix of size $n_{players} \times n_{players}$. Each $(i,j)^{\text{th}}$ entry of this square matrix denotes the number of chips belonging to player $j$ that are currently held by player $i$. 
Therefore, the term $n_{players}^2$ keeps track of the chips each player currently holds.

% $\text{num\_players}^2$. This term keeps track of the chips each player currently holds.

% Since each player can potentially hold chips that originated from any of the other players, we need a matrix-like structure with $\text{num\_players} \times \text{num\_players}$ entries to represent the state of chips per player.

% Each entry $(i, j)$ in this structure denotes the number of chips player $i$ holds from player $j$.

% The term $\text{num\_players}^2$ represents chip ownership across all players. 

% \subsubsection{Eliminated Chips}
% This term represents the chips belonging to players who have been eliminated, computed by $$n_{players}$$.
% As each player can potentially be eliminated during the game, the status of each player's chips must be tracked, and a single entry for each player is sufficient to mark their chips as dead.

% $\text{num\_players}$. 

% \subsubsection{Eliminated Chips}
% This term, of size $$n_{players}$$, tracks chips belonging to eliminated players. Since any player may be removed during the game, a single entry per player is sufficient to indicate whether their chips are considered dead.

\subsubsection{Eliminated Chips and Current Player}
% The \emph{Eliminated Chips} component, of size $$n_{players}$$, records whether a player’s chips are dead due to elimination, with one entry per player. 
% As each player can potentially be eliminated during the game, the status of each player's chips must be tracked, and a single entry for each player is sufficient to mark their chips as dead.
The \emph{eliminated chips} component, of size $n_{players}$, records whether each player’s chips are dead due to elimination. Since any player may be eliminated during the game, a single entry per player is sufficient to track this status. 
The \emph{current player} component, also of size $n_{players}$, is represented as a one-hot vector indicating which player is active at timestep $t$.

% \subsubsection{Current Player}
% xxx

\subsubsection{Game Phase}
% A vector of length $4$, representing the current phase or state of the game. There are four possible game states:
This component is a one-hot vector of length $4$ that encodes the current decision phase of the game: \textit{choose\_pile} (choosing a pile from which to play), 
\textit{choose\_chip} (selecting a chip to place in the chosen pile), 
\textit{choose\_next\_player} (designating the next player), 
or \textit{eliminate\_chip} (eliminating a chip from play).

%     \begin{itemize}
%         \item \texttt{choose\_pile}: The phase where the agent selects a pile from which to play.
%         \item \texttt{choose\_chip}: The phase where the agent selects a chip to use in the chosen pile.
%         \item \texttt{choose\_next\_player}: The phase where the agent selects the next player in the sequence.
%         \item \texttt{eliminate\_chip}: The phase where the agent eliminates a chip, marking it as no longer in play.
%     \end{itemize}
% Each of these states is represented by a unique one-hot encoding within the 4-element vector, indicating the current phase of the game.

% \subsubsection{Step Count}
% A scalar value indicating the number of steps taken in the current episode.
%     \begin{itemize}
%         \item This component provides a running count of the steps, which can help track the agent’s efficiency in gameplay. It may also be used in reward calculation to incentivize faster completion of tasks.
%     \end{itemize}
% \end{itemize}

\subsubsection{Step Count}
This component is a scalar that records the number of steps taken in the current episode. It serves both as a measure of gameplay efficiency and as a signal for reward shaping, where shorter episodes are incentivized.

% \noindent
Thus, the overall state representation or observation space size ($obs\_size$) is given by: 
\begin{equation}
\begin{split}
obs\_size = 
% \texttt{total\_obs\_size} = ~~~~~~~~~~~~~~~~~~~~~~~~~~~~~~~~~~~~~\\
(n_{rows} \times n_{players} \times n_{max\_pile}) \\
+~ n_{players}^2 ~+~ (2 \times n_{players}) + 4 + 1
\end{split}
\end{equation}

\subsection{Action Space}
The action $a^{<t>} \in {\mathcal{A}}$ taken by an agent at timestep $t$ is chosen from the action space $\mathcal{A}$ is discrete and consists of ten possible actions: 
% \begin{equation}
${\mathcal{A}} = \{ \textbf{\textscripta}_0 , \textbf{\textscripta}_1 , \ldots, \textbf{\textscripta}_9 \}$; 
% \end{equation}
where each action \textbf{\textscripta}$_i$ corresponds directly to a legal move defined by the game rules (refer to Section \ref{sec:rules}), and the action space can be grouped into two categories:

\begin{itemize}
    % \item Actions (\textbf{\textscripta}$_0$ – \textbf{\textscripta}$_5$) correspond to the selection of rows and are valid only in \emph{choose\_pile} phase. Each action in this range selects one of the six rows available. Up to six active piles can exist on the board at any given time. Each of the first six actions corresponds to selecting one of these piles. If fewer than six piles are available, the invalid options are dynamically masked during play.

\item  
Actions \textbf{\textscripta}$_0$ – \textbf{\textscripta}$_5$ correspond to pile selection and are valid only during the \emph{choose\_pile} phase. Each action selects one of up to six active piles on the board, with invalid choices dynamically masked when fewer than six piles are available.

\item 
Actions \textbf{\textscripta}$_6$ – \textbf{\textscripta}$_9$ correspond to player/color-based decisions and are valid during the \textit{choose\_chip}, \textit{choose\_next\_player}, and \textit{eliminate\_chip} phases. Each action refers to one of the four players (or equivalently, chip colors), enabling the agent to select a chip to play, designate the next player, or eliminate a chip, depending on the current game phase.

    % \item \emph{Chip Selection (\textbf{\textscripta}$_6$ – \textbf{\textscripta}$_9$):}  
    % The remaining four actions represent the four possible chip colors in the game. Once a pile has been chosen, the agent must select a chip color to play, and these actions specify that choice.
\end{itemize}

% This design yields a fixed-size action space of ten actions, divided into six pile-selection and four player/color-selection moves. At any given phase, only the relevant subset of actions is valid, while invalid choices are dynamically masked. Maintaining a constant action space simplifies the network output layer, whereas phase-dependent masking ensures compliance with the game’s sequential rules.

% \noindent 
This design yields a fixed-size action space while ensuring compatibility with the sequential decision-making structure of the game. At each phase, only a subset of actions is legal, and invalid actions are masked to maintain consistency during training. Keeping the action space size constant simplifies the network architecture, while action masking preserves adherence to game rules.

\begin{figure*}[t]
    \centering
\scriptsize 
(a) \includegraphics[width=0.45\linewidth]{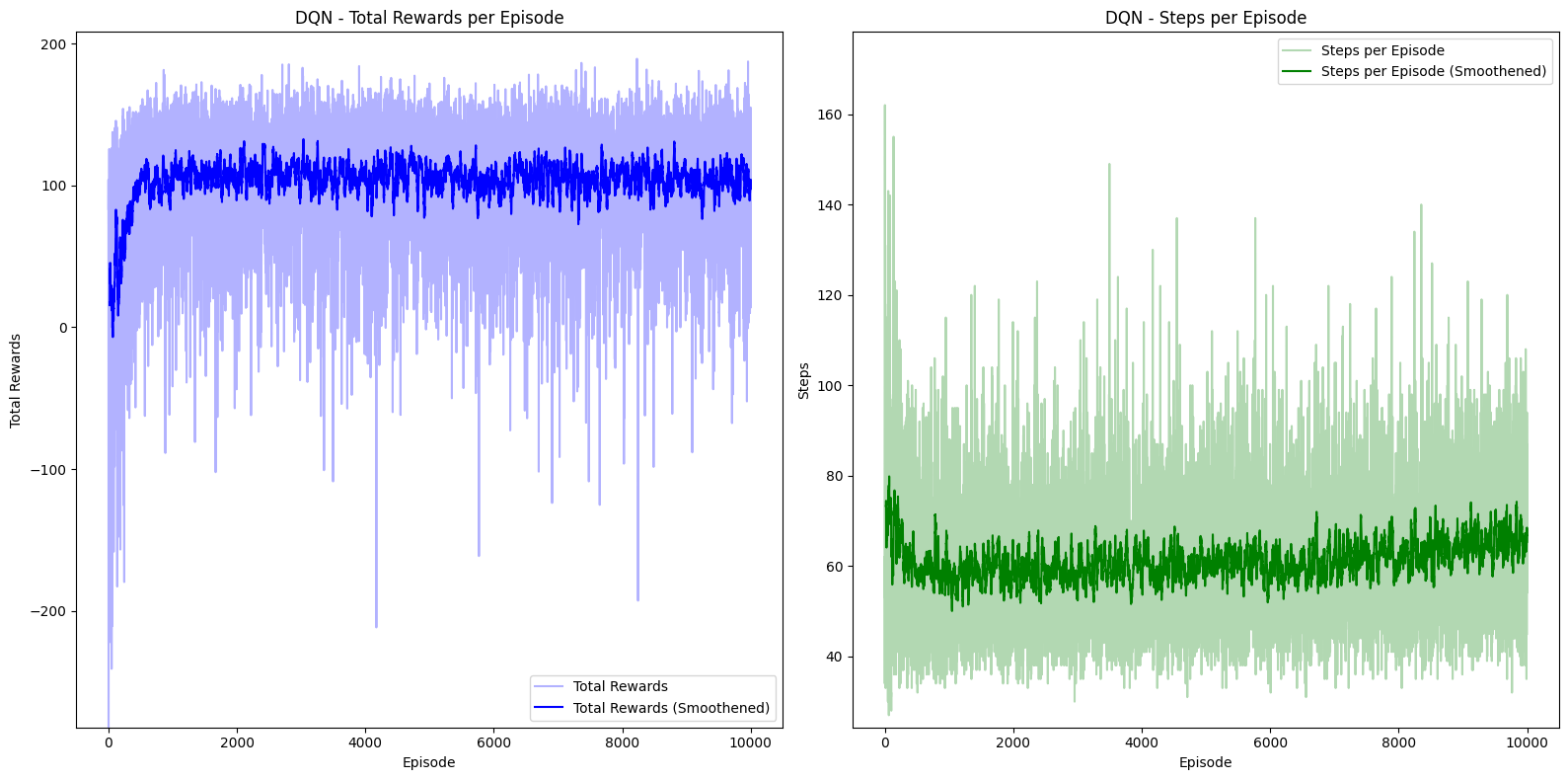}
(b) \includegraphics[width=0.45\linewidth]{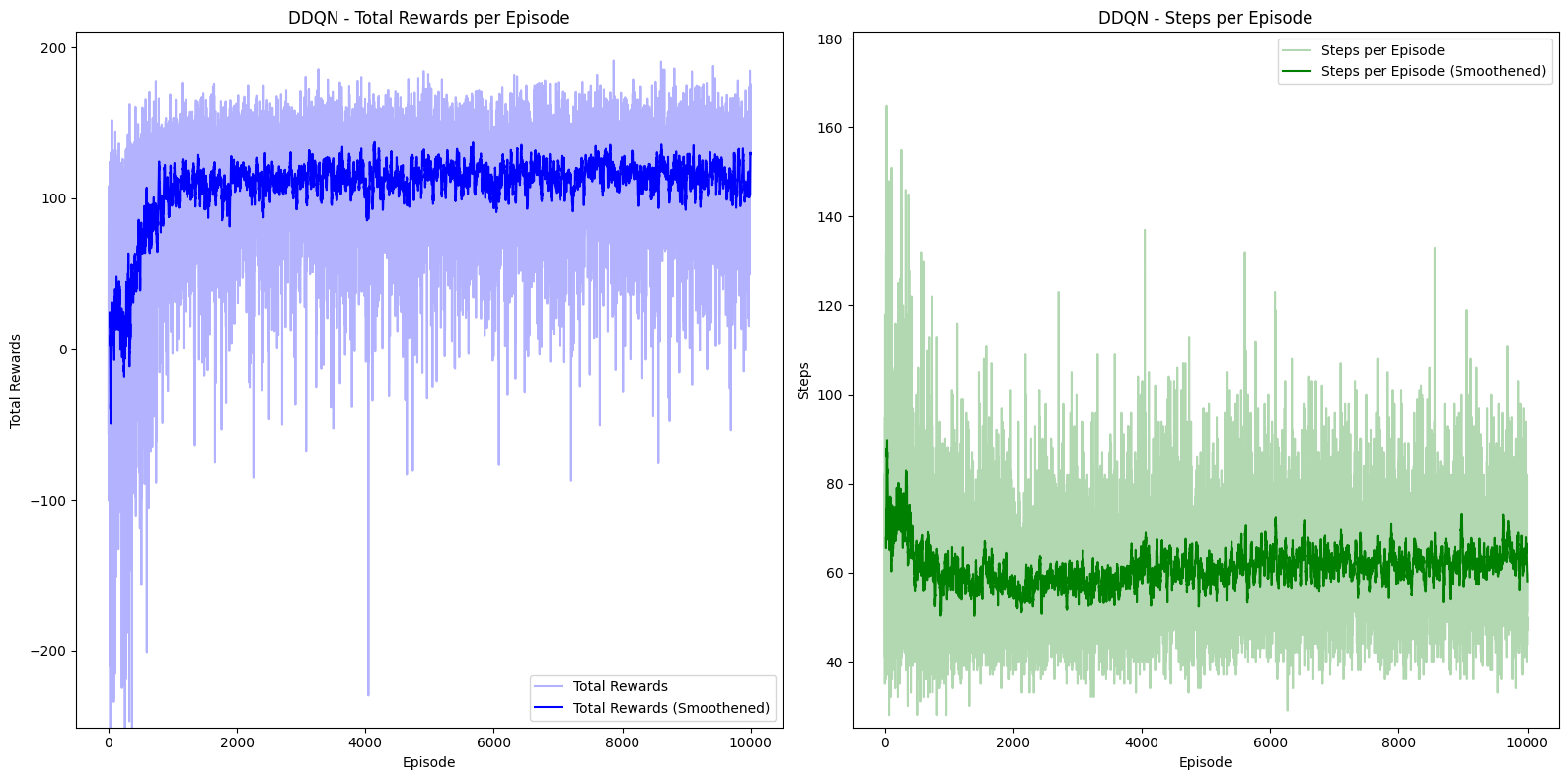}\\
(c) \includegraphics[width=0.45\linewidth]{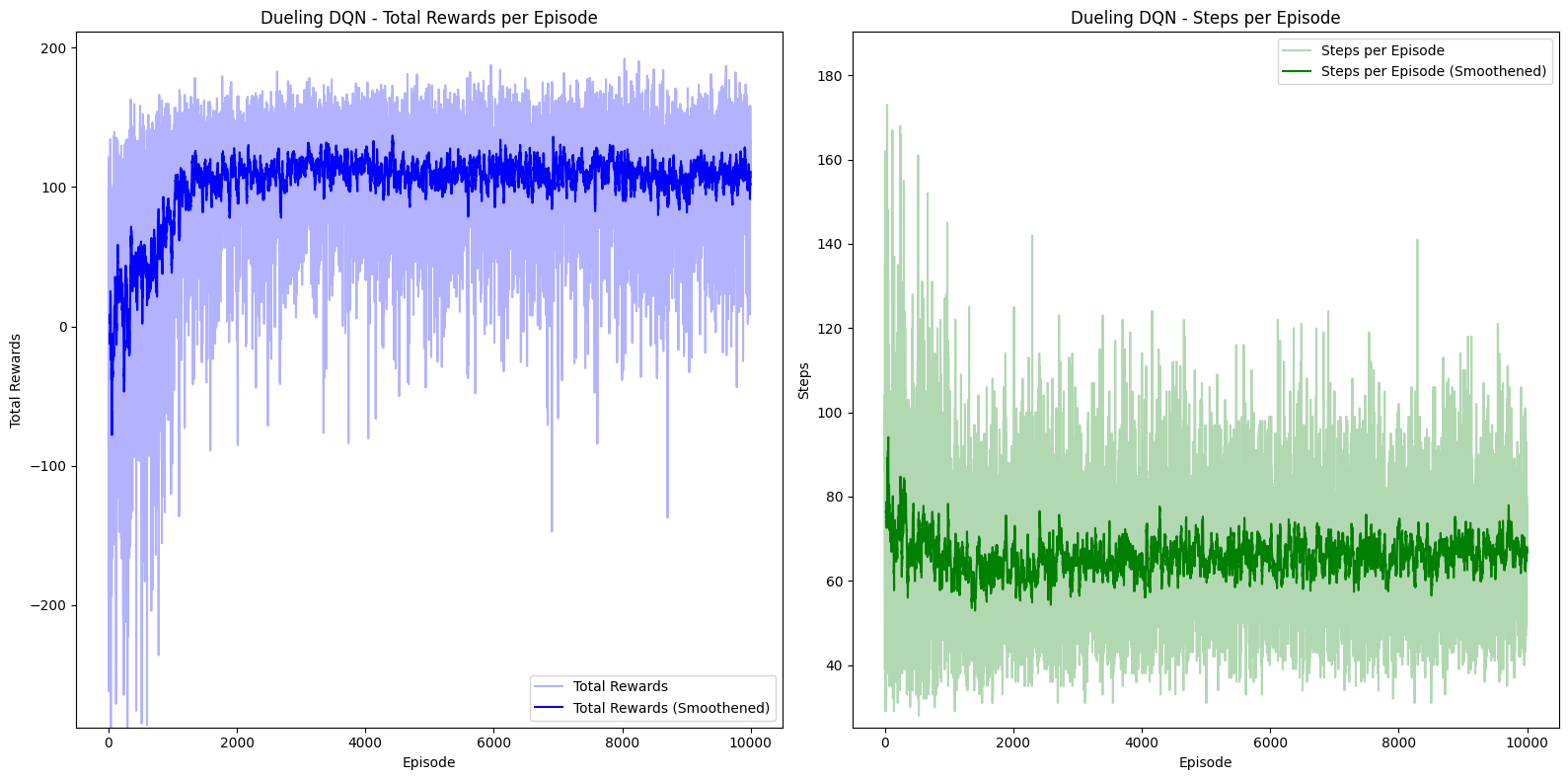}
(d) \includegraphics[width=0.45\linewidth]{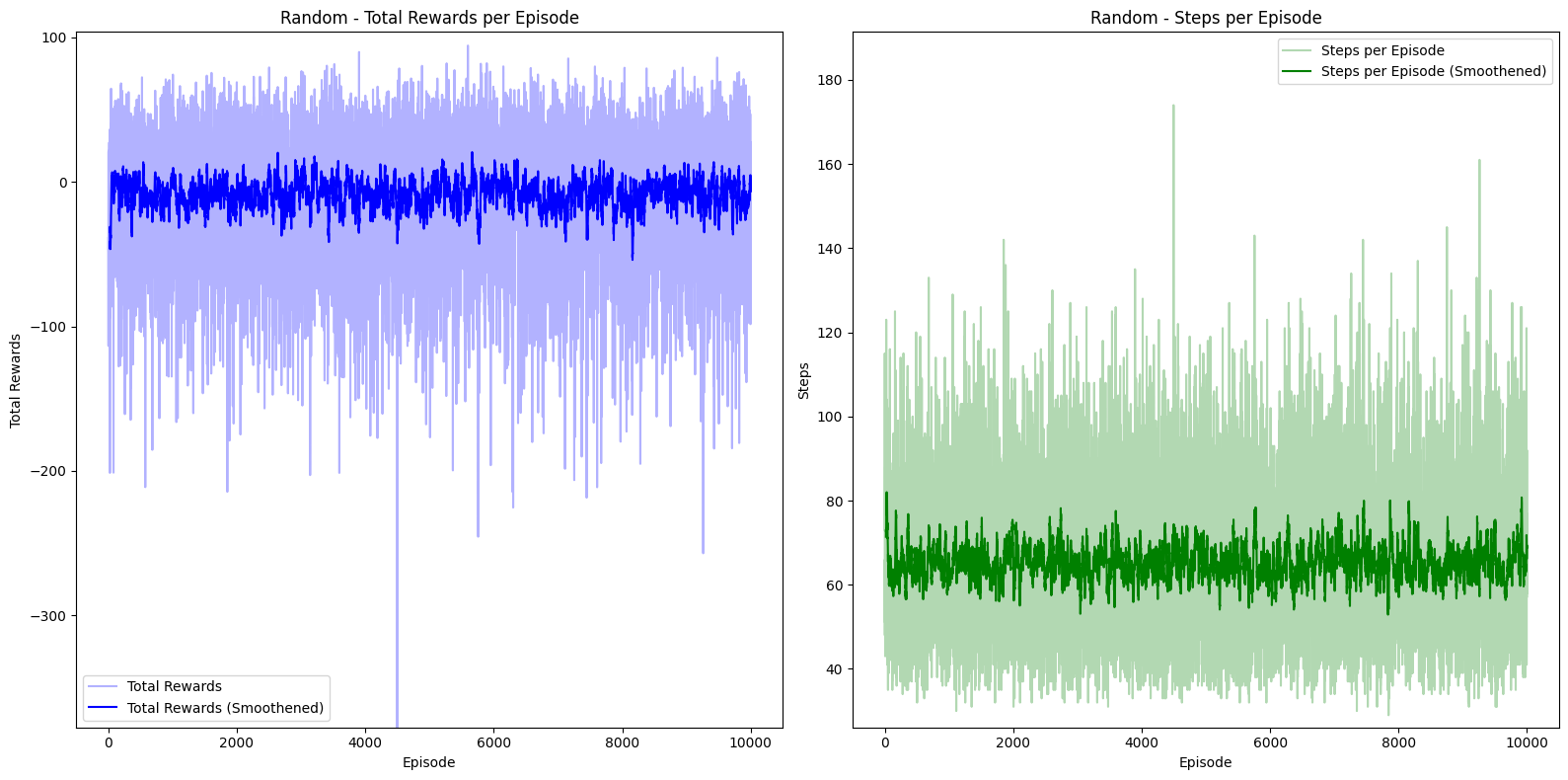}
% \caption{Performance comparison of different agents. Each subfigure depicts reward trends across episodes.}
\caption{Performance comparison of different agents:
(a) DQN: reward vs. episodes,
(b) DQN: steps vs. episodes,
(c) DDQN: reward vs. episodes,
(d) DDQN: steps vs. episodes,
(e) Dueling DQN: reward vs. episodes,
(f) Dueling DQN: steps vs. episodes.}

    \label{fig:agent_performance}
\end{figure*}

\subsection{Reward Design}
The reward signal serves as the primary feedback mechanism that drives learning. Positive rewards encourage desirable behaviors such as legal and efficient play, while negative rewards penalize undesirable outcomes such as illegal moves or prolonged gameplay. By accumulating rewards across episodes, the agent is guided toward policies that maximize long-term performance rather than short-term gains, thereby capturing the sequential and strategic nature of SLS. 

At each timestep $t$, the reward $r_t$ is shaped as follows: 
\begin{itemize}
    \item 
    \emph{illegal actions} incur a fixed negative reward ($-\wp \in \mathbb{R}^-$), penalizing rule violations; 
    \item 
    \emph{legal actions} provide a positive reward of up to $+\wp \in \mathbb{R}^+$, which decays with the number of steps and chips involved. Each valid move is initially rewarded, but the value decreases dynamically to promote efficiency. 
\end{itemize}
Since the number of chips $n_c$ allocated to each player at the start of the game is fixed ($n_c=5$, see Section~\ref{sec:rules}), the reward adjustment primarily depends on the step count. The reward $r_t$ at step $t$ is defined as: 
\begin{equation}
    r_t = \text{min} \left( \wp ~,~ \frac{\wp}{ ( \frac{\alpha}{n_c})~.~ t}  \right);
\end{equation}
%
% $r_t = \text{min} \left( \wp ~,~ \frac{\wp}{ ( \frac{\alpha}{n_c})~.~ t}  \right)$; 
%
where, $0 < \alpha \leq 1$ is a hyperparameter controlling the decay rate. 
% In our experiments, we set $\alpha = 0.3$, $\wp=5$, decided empirically. 
The hyperparameters $\alpha = 0.3$ and $\wp = 5$ were selected empirically through exploratory experiments. 
This dynamic reward shaping incentivizes agents to complete tasks using fewer steps. As $t$ increases, the effective reward decreases, discouraging unnecessary actions and ensuring that learning aligns with competitive, rule-compliant play.

% \begin{equation}
%     r_t = \text{min} \left( \wp ~,~ \frac{\wp . n_c}{ \alpha ~.~ t}  \right)
% \end{equation}
    
    % \[
    % r_t = \min \left( 5, \frac{5}{\alpha \cdot \text{steps}} \right),
    % \]

    % where $\alpha$ normalizes for chip count and pacing.

\section{Experiments and Discussion}
\label{4sec:result}

\subsection{Cumulative Learning Framework}
We employ a centralized cumulative learning setup \cite{curriculum}, where all four player agents share a common learning network, such as DQN, and update a shared replay buffer. 
This improves sample efficiency by exposing the network to diverse transitions while maintaining decentralized execution at inference time. 
Such a design highlights how SLS stresses long-horizon reasoning even when agents have identical training signals.

% \subsection{Agent Training Workflow}

% The overall agent training process is summarized in Algorithm~\ref{alg:multiagent}. At each episode, the environment is initialized and agents act according to the current policy. Valid actions are sampled based on the game phase. Transitions are stored in the shared buffer, and gradient updates are periodically applied to the network. Exploration decays over time, and the target network is periodically updated.

% \newpage

% \subsection{Training Workflow}
% The training loop proceeds as follows:
% \begin{enumerate}
%     \item Initialize environment and reset game state.
%     \item On each turn, the current agent selects an action using an $\epsilon$-greedy policy.
%     \item Execute the action, record transition $(s_t,a_t,r_t,s_{t+1})$ in the shared buffer.
%     \item Periodically update the Q-network with minibatch sampling and synchronize the target network.
%     \item Reduce exploration rate over time until convergence.
% \end{enumerate}
% This workflow provides a fair and reproducible baseline for evaluating MARL performance in SLS.

% \input{4_result}

\subsection{Training Configuration and Hyperparameters}
To evaluate the proposed SLS environment, we trained and compared three classical DRL agents, 
DQN, DDQN, Dueling DQN, against a random baseline. All agents shared a centralized replay buffer and were implemented using TensorFlow with the Adam optimizer. The Q-network comprised two fully connected hidden layers with $64$ neurons each and ReLU activation, followed by a linear output layer generating Q-values for each action. The model was trained for 10000 episodes using MSE loss. 
The key hyperparameters were empirically set as follows: 
discount factor ($\gamma$) = 0.95, 
initial exploration rate ($\epsilon_0$) = 1.0, 
exploration decay rate ($\epsilon_{\text{decay}}$) = 0.995, 
minimum exploration rate ($\epsilon_{\min}$) = 0.01, 
learning rate = 0.001, and 
batch size = 64.

\subsection{Quantitative Performance}
Table~\ref{tab:performance} summarizes the agents’ cumulative rewards and average step counts per episode. Each reported value is computed over multiple runs to ensure stability and reproducibility.

\begin{table}[t]
\centering
\caption{Agent Performance Metrics}
\label{tab:performance}
\begin{adjustbox}{width=\linewidth}
\begin{tabular}{@{}lcccc@{}}
\toprule
\textbf{Agent} & \textbf{Reward} & \textbf{Reward Range} & \textbf{Step} & \textbf{Step Range} \\
 & \textbf{(mean $\pm$ stdev)} & \textbf{[min, max]} & \textbf{(mean $\pm$ stdev)} & \textbf{[min, max]} \\
\midrule
DQN & 103.40 $\pm$ 42.31 & [--313.45, 189.24] & 61.16 $\pm$ 14.51 & [27, 162] \\
DDQN & {108.44} $\pm$ 44.95 & [--279.13, 191.38] & 61.23 $\pm$ 14.18 & [28, 165] \\
Dueling DQN & 102.06 $\pm$ 49.62 & [--319.76, 192.09] & 65.92 $\pm$ 15.94 & [28, 173] \\
Random & --8.78 $\pm$ 43.52 & [--419.26, 94.19] & 65.24 $\pm$ 17.76 & [29, 174] \\
\bottomrule
\end{tabular}
\end{adjustbox}
\end{table}

All DRL agents consistently outperformed the random baseline, achieving roughly half of the theoretical maximum reward ($\approx200$). Among them, DDQN achieved the most stable convergence and the highest mean reward, validating the benefit of double estimation in mitigating Q-value overestimation for long-horizon games.

\subsection{Learning Dynamics}
Fig.~\ref{fig:agent_performance} illustrates the episode-wise reward trends. During the early training phase ($<500$ episodes), agents exhibited substantial reward variance due to exploration. Beyond approximately $2000$ episodes, all DRL agents displayed smoother convergence, with mean rewards stabilizing near 100--120. Occasional dips correspond to exploration bursts or misestimation of legal action sequences. Despite the incorporation of dynamic action masking, all agents still committed occasional illegal moves, reflecting the complexity of SLS’s combinatorial action space and the delayed-reward dependencies.

% \begin{figure}[h]
%     \centering
%     \includegraphics[width=0.9\linewidth]{reward_trends.png}
%     \caption{Learning dynamics of DQN, DDQN, and Dueling DQN agents across 2000 training episodes. Rewards are smoothed using a 20-episode moving average.}
%     \label{fig:learning_trends}
% \end{figure}

% \subsection{Discussion and Insights}
% The empirical results demonstrate that classical value-based methods can capture the fundamental rules and partial strategic structure of SLS. However, the wide variance in reward distributions (Table~\ref{tab:performance}) indicates sensitivity to initial states and exploration schedules. Context-dependent actions such as “choose next player” require long-horizon reasoning, which remains challenging for purely value-based learners. Consequently, agents tend to exhibit reactive rather than anticipatory strategies.

% These findings underscore both the promise and the limitations of classical DRL in negotiation-aware multi-agent games. Future work may explore (i) actor–critic algorithms such as PPO or A3C for improved sample efficiency, (ii) hierarchical or rule-augmented critics to enforce legality, and (iii) coalition-aware policy conditioning to better capture the social dynamics inherent to SLS. Together, these directions could move toward agents capable of not only legal and efficient play but also strategic negotiation and deception—core aspects of the original game design.

\subsection{Discussion and Insights}
The results show that classical value-based methods can learn the core rules and partial strategies of \textit{So Long Sucker} (SLS), achieving stable yet suboptimal performance. The high variance in rewards (Table~\ref{tab:performance}) reflects sensitivity to initialization and exploration, while context-dependent actions expose the limitations of short-horizon value estimation. Consequently, the agents often act reactively rather than strategically. 
These findings highlight both the potential and constraints of classical DRL in negotiation-aware multi-agent games. 
Future directions involve integrating actor–critic and coalition-aware frameworks to strengthen long-term reasoning, enforce rule compliance, and enable adaptive strategic behavior, ultimately advancing agents toward negotiation/ alliance/ deception-capable play, the core essence of SLS.

% Future directions include exploring actor–critic and coalition-aware frameworks to enhance long-term reasoning, legality enforcement, and adaptive strategy formation, moving toward agents capable of negotiation, alliance, and deception—the essence of SLS.

\section{Conclusion}
\label{5sec:conclusion}

This paper introduced a computational framework for the strategy game SLS, establishing it as a negotiation-aware benchmark for MARL. Using classical DRL algorithms, DQN, DDQN, and Dueling DQN, we showed that agents can learn legal and moderately strategic gameplay, achieving about half of the maximum attainable reward and outperforming random baselines. While these agents captured fundamental rules, they struggled with long-horizon reasoning and coalition dynamics, underscoring the game’s inherent complexity. 
Future work will extend this framework with actor–critic and coalition-aware approaches to enable richer strategic behavior.

\bibliographystyle{IEEEtran}  
\bibliography{ref.bib} 

% \fi 

\end{document}